\documentclass[10pt,twocolumn,letterpaper]{article}

\usepackage{iccv}              

\definecolor{iccvblue}{rgb}{0.21,0.49,0.74}
\usepackage[pagebackref,breaklinks,colorlinks,allcolors=iccvblue]{hyperref}
\usepackage[accsupp]{axessibility}

\usepackage{amsmath}
\usepackage{tabularx}
\usepackage{multicol}
\usepackage{multirow}
\usepackage{adjustbox}
\usepackage{booktabs}
\usepackage{xspace}
\usepackage{hyperref}
\usepackage{algorithm}
\usepackage{algpseudocode}
\usepackage{array}
\usepackage{xcolor}

\usepackage{enumitem}
\usepackage{orcidlink}

\def\paperID{15} 
\def\confName{ICCV}
\def\confYear{2025}

\newcommand{\methname}{RAD\xspace}

\title{Robust Anomaly Detection in Industrial Environments via Meta-Learning}

\author{Muhammad Aqeel$^1$\orcidlink{0009-0000-5095-605X}\and Shakiba Sharifi$^1$\orcidlink{0009-0008-6309-635X} \and Marco Cristani$^{1,2}$\orcidlink{0000-0002-0523-6042} \and Francesco Setti$^{1,2}$\orcidlink{0000-0002-0015-5534} \\
$^1$ Dept. of Engineering for Innovation Medicine, University of Verona\\
Strada le Grazie 15, Verona, Italy\\
$^2$ Qualyco S.r.l., Strada le Grazie 15, Verona, Italy\\
{\tt\small Contact author: muhammad.aqeel@univr.it}
}

\begin{document}
\maketitle
%
\begin{abstract}
Anomaly detection is fundamental for ensuring quality control and operational efficiency in industrial environments, yet conventional approaches face significant challenges when training data contains mislabeled samples—a common occurrence in real-world scenarios. This paper presents \methname, a robust anomaly detection framework that integrates Normalizing Flows with Model-Agnostic Meta-Learning to address the critical challenge of label noise in industrial settings. Our approach employs a bi-level optimization strategy where meta-learning enables rapid adaptation to varying noise conditions, while uncertainty quantification guides adaptive L2 regularization to maintain model stability. The framework incorporates multiscale feature processing through pretrained feature extractors and leverages the precise likelihood estimation capabilities of Normalizing Flows for robust anomaly scoring. Comprehensive evaluation on MVTec-AD and KSDD2 datasets demonstrates superior performance, achieving I-AUROC scores of 95.4\% and 94.6\% respectively under clean conditions, while maintaining robust detection capabilities above 86.8\% and 92.1\% even when 50\% of training samples are mislabeled. The results highlight \methname's exceptional resilience to noisy training conditions and its ability to detect subtle anomalies across diverse industrial scenarios, making it a practical solution for real-world anomaly detection applications where perfect data curation is challenging.
%
\end{abstract}

\section{Introduction}
\label{sec:intro}

\begin{figure}[t!]
    \centering
    \includegraphics[height=8cm, width=\linewidth, keepaspectratio]{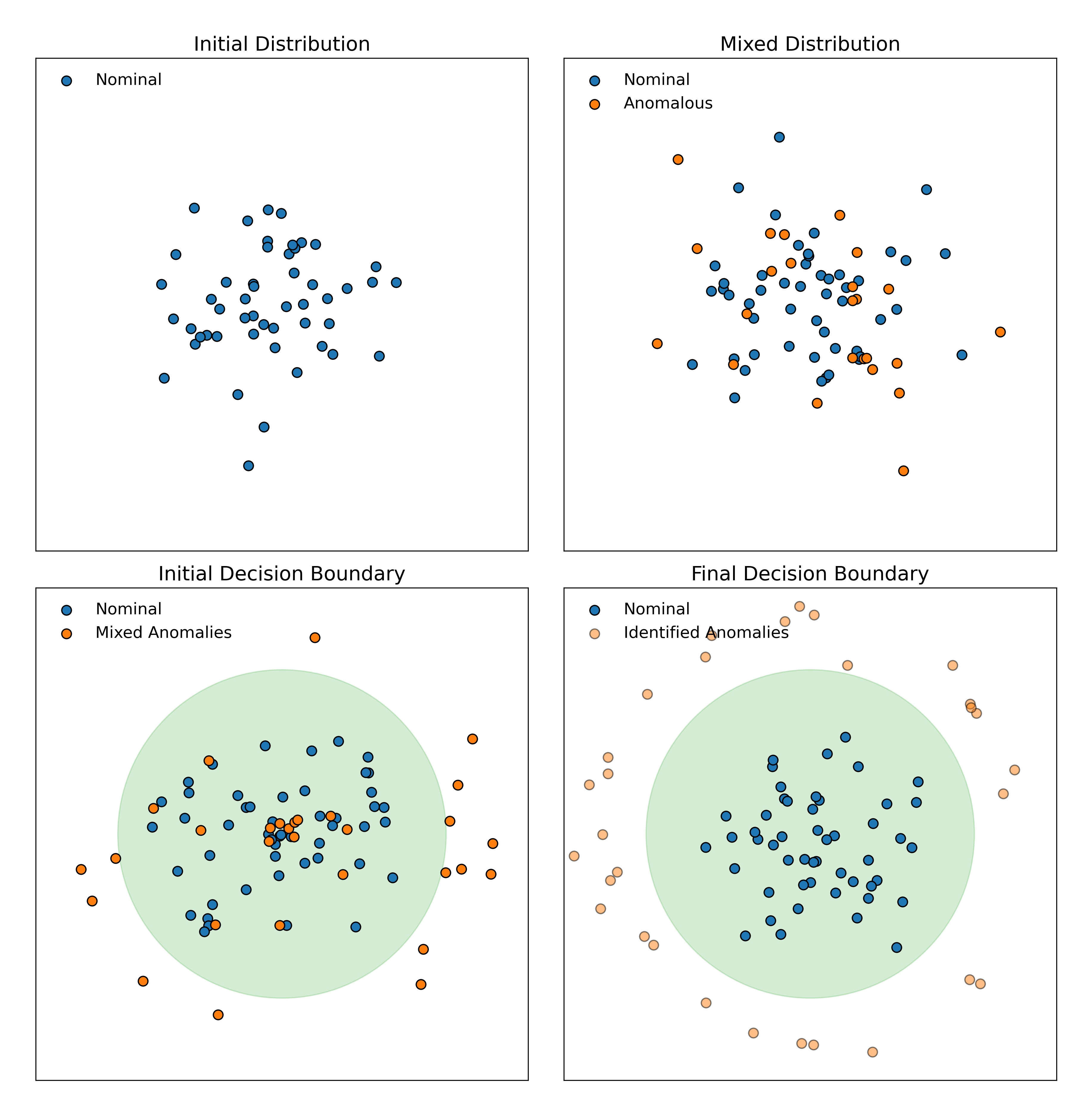}
    \caption{Conceptual demonstration of model robustness in anomaly detection. Data points represent feature space distributions where blue indicates nominal samples and orange indicates anomalous samples. The top row shows that while traditional models struggle to distinguish anomalies that behave like nominal samples (right) from pure nominal data (left), our model maintains clear decision boundaries. The bottom row illustrates how our approach successfully identifies and separates boundary-crossing anomalies (left), resulting in robust anomaly separation (right) even when anomalies closely mimic nominal behavior.}
    \label{fig:training}
\end{figure}

In the domain of industrial machinery, effective anomaly detection is pivotal to ensuring operational integrity and maximizing system efficiency. Industrial systems continuously produce complex, high-dimensional data streams reflecting equipment performance across diverse operational states~\cite{wang2019industrial}. Anomalies in this context refer to deviations from normal operational patterns that may indicate defects, malfunctions, or quality issues. Identifying these anomalous patterns is essential, as these deviations often signal incipient failures, safety risks, or inefficiencies that, if left unaddressed, may lead to costly disruptions or hazards~\cite{liu2024deep}. However, detecting subtle and context-sensitive anomalies embedded within high-dimensional, non-stationary data distributions represents a substantial technical challenge, particularly when training data contains noise in the form of mislabeled samples. Figure~\ref{fig:training} illustrates this challenge through a conceptual visualization in feature space, where data points represent distributions of nominal (blue) and anomalous (orange) samples. Traditional models struggle to maintain clear decision boundaries when anomalies closely mimic nominal behavior, leading to misclassification.

Anomaly detection is challenging due to the rarity, variability, and context-dependent nature of anomalies. Traditional statistical and machine learning methods rely on fixed data distribution assumptions~\cite{aqeel2025CoMet, aggarwal2017introduction,chandola2009anomaly}. While foundational, these approaches struggle in industrial settings with complex, non-stationary data and intricate dependencies~\cite{zhang2021understanding}. The scarcity of anomalies and the high cost of labeled data further complicate detection~\cite{pang2021deep}. 

Deep learning has revolutionized image anomaly detection, with CNNs and other architectures excelling at anomaly detection and localization in high-dimensional image data~\cite{bergmann2020uninformed,chow2020anomaly}. Deep generative models like VAEs, GANs, and NFs have proven especially effective by learning normal data distributions to identify anomalies~\cite{schlegl2017unsupervised, akcay2019ganomaly,rezende2015variational}. These unsupervised approaches eliminate the need for labeled anomaly datasets, making them ideal for industrial applications~\cite{ruff2021unifying,pang2021toward}.

However, state-of-the-art methods face challenges in robustness and generalization. Issues such as boundary sensitivity --where legitimate yet unusual normal data points near the boundary between nominal and anomalous distributions are misclassified-- persist and lead to the generation of false negatives~\cite{zhang2021understanding,pang2021deep}. Additionally, real-world industrial data often contains label noise, where some anomalous samples may be incorrectly labeled as nominal during training, further degrading model performance. Addressing these limitations is critical for applications where missed detections and false alarms can have significant operational and economic consequences~\cite{pang2021toward}.

This research introduces \emph{Robust Anomaly Detection (\methname)}, a robust framework for anomaly detection in images that specifically addresses noise robustness --the ability to maintain high performance when training data contains mislabeled samples. Our approach addresses the boundary sensitivity issue by creating more resilient decision boundaries that successfully separate anomalies regardless of their proximity to nominal data. \methname integrates meta-learning, iterative refinement, and L2 regularization to address the inherent challenges of the task. The framework is designed to improve the model's capacity to generalize across diverse and dynamic data distributions while preserving sensitivity to subtle and complex anomaly patterns. By integrating the capabilities of deep generative modeling with L2 regularization to address overfitting, \methname achieves an optimal balance between precision and generalization, effectively minimizing false positives and false negatives even under noisy training conditions.

The main contributions of our paper can be summarized as follows: 
\begin{itemize}
\item We propose \methname, a novel anomaly detection framework that leverages Normalizing Flows and meta-learning to adaptively refine decision boundaries and handle complex industrial data distributions under noisy training conditions.
\item \methname incorporates multiscale feature processing and scale-translation networks, enabling precise detection of subtle anomalies across varying conditions while maintaining robustness to label noise.
\item Comprehensive experiments on MVTec-AD~\cite{bergmann2019mvtec} and KSDD2~\cite{Bozic2021COMIND} demonstrate the state-of-the-art performance of \methname, maintaining AUROC scores above 86.8
\end{itemize}

\begin{figure*}[t!]
    \centering
    \includegraphics[width=1.0\linewidth,trim={0 .0cm 0 0},clip]{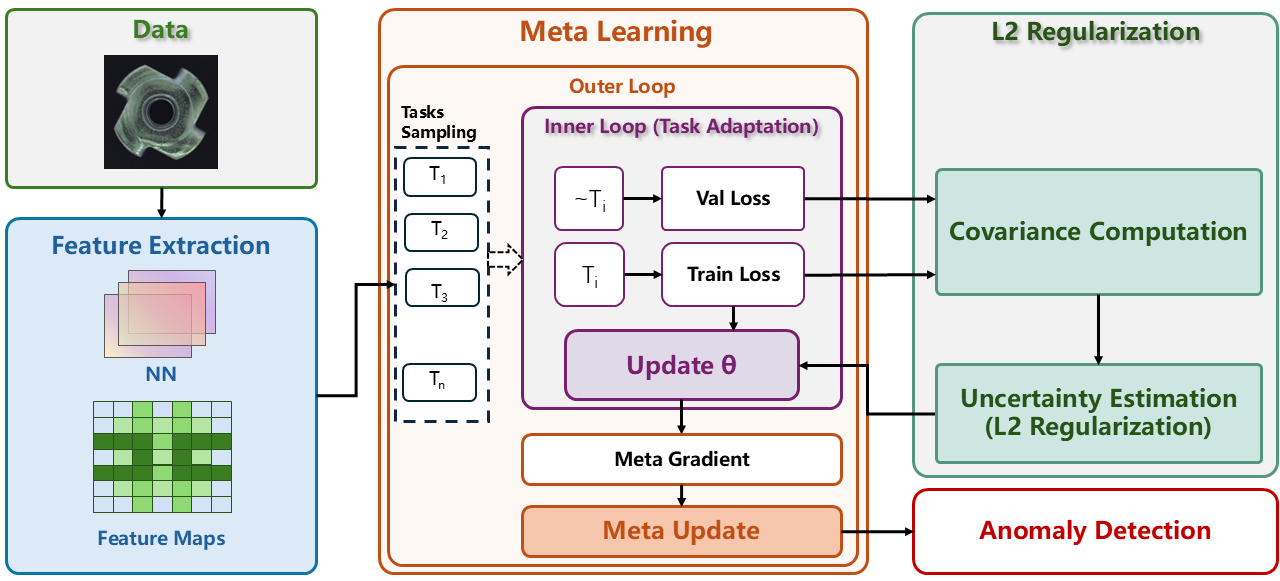}
    \caption{\methname pipeline. Input images are processed through a pretrained feature extractor before being fed into the Normalizing Flow module for density estimation. The meta-learning framework employs bi-level optimization with inner and outer loops to adaptively refine model parameters, while uncertainty quantification via covariance analysis guides the application of L2 regularization for enhanced robustness.}
    \label{fig:pipeline}
\end{figure*}
\section{Related Work}
\label{sec:realtedwork}

Anomaly detection in images has evolved significantly from traditional statistical methods to sophisticated deep learning-based frameworks. Traditional approaches, such as Gaussian-based models, Support Vector Machines (SVMs), and decision trees, have laid a foundational understanding of anomaly detection but often struggle with the high-dimensional nature of image data in industrial settings~\cite{aqeel2025CoMet, chandola2009anomaly,hodge2004survey}. These limitations have driven a transition towards advanced neural network architectures that can scale and generalize effectively.

Deep learning models such as Convolutional Neural Networks (CNNs) and Variational Autoencoders (VAEs) have emerged as prominent tools for anomaly detection. CNNs excel at extracting hierarchical features from raw images, enabling superior performance in tasks such as defect detection~\cite{bergmann2019mvtec}. Variational Autoencoders and Generative Adversarial Networks (GANs) extend these capabilities by modeling data distributions and detecting anomalies through reconstruction errors or deviations from generated norms~\cite{schlegl2017unsupervised,ruff2021unifying}. Despite their strengths, these methods are sensitive to the quality of training data and prone to overfitting in high-noise environments.

Unsupervised methods such as Autoencoders and Normalizing Flows (NFs) offer solutions by learning normal data patterns without requiring labeled anomaly data. NFs are particularly effective in modeling complex probability distributions by transforming them into simpler and more tractable forms, enabling precise identification of subtle anomalies~\cite{rezende2015variational}. However, NF training can be computationally intensive and sensitive to hyperparameter configurations.

Recent advancements in meta-learning and self-supervised techniques have significantly improved anomaly detection by enhancing adaptability to dynamic data distributions and operational conditions~\cite{finn2017model, hospedales2021meta}. Confident Learning (CL) has further contributed to robustness by identifying mislabeled samples and reducing false positives~\cite{northcutt2021confident}. Several contemporary approaches have leveraged iterative refinement strategies for more precise anomaly detection in industrial inspection~\cite{aqeel2025meta, Aqeel_2025}, while self-supervised learning techniques have demonstrated strong potential in detecting surface defects with minimal labeled data~\cite{aqeel2024self}. Additionally, normalizing flow-based methods have been employed to enhance semi-supervised defect detection, effectively distinguishing subtle anomalies from nominal patterns~\cite{RudWan2021}.

Building on these advancements, our research integrates meta-learning, label refinement techniques, and progressive optimization strategies to develop a robust anomaly detection framework. This framework ensures adaptability and precision in detecting anomalies in complex and dynamic industrial environments by addressing challenges such as noise, variability, and near-boundary ambiguity.


\section{\methname Pipeline}
\label{sec:method}

This section presents the comprehensive methodology for robust anomaly detection, integrating Normalizing Flows (NF), Model-Agnostic Meta-Learning (MAML), and uncertainty-guided $L_2$ regularization. The framework is engineered to deliver accurate anomaly detection, cross-dataset adaptability, and robustness against noisy training conditions. Figure~\ref{fig:pipeline} illustrates the complete system architecture, while Algorithm~\ref{algo:meta_learning} formalizes the progressive optimization framework.

\subsection{Feature Extraction and Transformation}
\label{ssec:Feature}

The methodology begins with extracting features using a pre-trained feature extractor inspired by previous works \cite{rudolph2023asymmetric,RudWan2021}. The extractor maps input images $x \in \mathcal{X}$ to a multi-scale feature space $u \in \mathcal{U}$. This representation captures fine-grained and coarse image details, which is crucial for distinguishing subtle anomalies from normal variations.

To transform the feature representation into a latent space $z$ with a Gaussian distribution $p(z)$, Normalizing Flows (NF) are employed. The NF model, parameterized by $\theta$, applies invertible transformations involving translation functions ($\tau$) and scale ($\sigma$) functions. 

This mapping is expressed as follows:

\begin{equation}
p_\theta(u) = p_\theta(z) \cdot \left| \det \frac{\partial z}{\partial u} \right|,
\end{equation}
where $\left| \det \frac{\partial z}{\partial u} \right|$ represents the determinant of the Jacobian matrix of the transformation. The invertibility of NF ensures an exact likelihood computation, enabling robust density estimation.

The NF model is trained by minimizing the negative log-likelihood of the features $u$:

\begin{equation}
\mathcal{L}_{\text{NF}}(u \mid \theta) = \| z \|_2^2 - \log \left| \det \frac{\partial z}{\partial u} \right|,
\end{equation}
where the first term, $\| z \|_2^2$, encourages features to cluster near $z = 0$ in the latent space, and the second term penalizes trivial transformations by incorporating the log-determinant of the Jacobian, thereby promoting meaningful feature mappings.

During inference, features with low likelihood are flagged as anomalies. To improve robustness, an anomaly score is computed by averaging negative log-likelihoods across multiple transformations $\mathcal{S}_i(x)$ such as rotations, translations, and flips:

\begin{equation}
a_\theta(x_i) = \mathbb{E}_{\mathcal{S}_i} \left[ -\log p\left(f_\theta\left(f_\phi\left(\mathcal{S}_i(x_i)\right)\right)\right) \right],
\end{equation}
where $f_\phi$ is the feature extractor and $f_\theta$ represents the NF model.

\subsection{Quantifying Uncertainty and L2 Regularization}

To handle noisy data and ensure model stability, the uncertainty is quantified using the determinant of the covariance matrix $\Sigma$, inspired by \cite{kendall2017uncertainties}. This matrix encapsulates variability in training and validation losses:
\begin{equation}
\Sigma =
\begin{bmatrix}
\text{Cov}(\mathcal{L}_{\text{train}}, \mathcal{L}_{\text{train}}) & \text{Cov}(\mathcal{L}_{\text{train}}, \mathcal{L}_{\text{val}}) \\
\text{Cov}(\mathcal{L}_{\text{val}}, \mathcal{L}_{\text{train}}) & \text{Cov}(\mathcal{L}_{\text{val}}, \mathcal{L}_{\text{val}})
\end{bmatrix}.
\end{equation}
The determinant $\det(\Sigma)$ serves as a scalar measure of overall uncertainty: 
High $\det(\Sigma)$ indicates significant variability in model performance across training and validation, suggesting higher uncertainty, while low values of $\det(\Sigma)$ reflects stable and consistent performance, indicating confidence in the model's learning process.

To ensure stability and generalization, $\mathcal{L}_2$ regularization is incorporated into the model. This term penalizes large parameter values, encouraging a simpler model that avoids overfitting. The regularized loss is defined as:
\begin{equation}
\mathcal{L}_{\mathrm{NF\_reg}}(\theta) = \mathcal{L}_{\mathrm{NF}} + \lambda \cdot \|\theta\|_2^2,
\end{equation}
where $\mathcal{L}_{\mathrm{NF}}$ is the original loss function of the Normalizing Flows model, $\|\theta\|_2^2$ is the squared $\mathcal{L}_2$-norm of the model parameters, and $\lambda$ is the regularization coefficient that controls the strength of the penalty.

By incorporating $\mathcal{L}_2$ regularization, the methodology ensures a balanced trade-off between model complexity and performance, improving robustness in noisy conditions.

\subsection{Meta-Learning}

To enhance adaptability in noisy industrial environments, we integrate Model-Agnostic Meta-Learning (MAML)~\cite{finn2017model} with our Normalizing Flow architecture. This integration enables the model to rapidly adapt decision boundaries when encountering diverse data distributions during training, which is critical for anomaly detection in dynamic industrial environments where data characteristics may vary significantly.

Our meta-learning framework operates through a bi-level optimization strategy designed specifically for normalizing flow-based anomaly detection. In the inner loop, task-specific parameters are updated using gradient descent to adapt to specific data distributions:

\begin{equation}
\theta' = \theta - \alpha \nabla_\theta \mathcal{L}_{\text{train}}(\theta),
\label{eq:innerloop}
\end{equation}

where $\alpha$ is the inner-loop learning rate. The outer loop optimizes the initial parameters to generalize across different noise conditions and data distributions:

\begin{equation}
\theta \leftarrow \theta - \beta \nabla_\theta \mathcal{L}_{\text{meta}}(\theta'),
\label{eq:outerloop}
\end{equation}

where $\beta$ is the meta-learning rate. The meta-objective function integrates the Normalizing Flow loss with uncertainty-guided adaptive L2 regularization:

\begin{equation}
\mathcal{L}_{\text{meta}}(\theta') = \sum_{i=1}^N \mathcal{L}_{\text{NF}}(x_i|\theta') + \lambda \|\theta'\|_2^2,
\end{equation}

where the regularization coefficient $\lambda$ is dynamically adjusted based on uncertainty quantification from the covariance matrix determinant. This formulation ensures model stability while maintaining sensitivity to genuine anomalies under varying noise conditions.

The bi-level optimization enables rapid adaptation to new distributions without catastrophic forgetting, while the uncertainty-guided regularization provides enhanced robustness. This framework facilitates improved boundary refinement through iterative optimization, addressing the specific challenges of industrial anomaly detection where both adaptability and stability are essential.







\subsection{Adaptive Refinement}
An adaptive refinement loop is implemented to continuously improve anomaly detection. Anomaly scores are refined using a threshold based on the interquartile range:

\begin{equation}
t = Q_3 + k \cdot (Q_3 - Q_1),
\label{eq:threshold}
\end{equation}
where $Q_1$ and $Q_3$ are the first and third quartile respectively, and $k$ adjusts the sensitivity of the threshold. This adaptive thresholding mechanism accounts for changes in anomaly score distributions, ensuring robust and consistent detection across diverse datasets.

\subsection{Bayesian Optimization for Hyperparameter Tuning}
Hyperparameters such as learning rates ($\alpha, \beta$) in equations \ref{eq:innerloop} and \ref{eq:outerloop} respectively, and the threshold adjustment factor ($k$) in equation \ref{eq:threshold} for anomaly detection play a critical role in the model's performance. Bayesian Optimization is employed to identify their optimal values. We use a Gaussian Process to model the relationship between hyperparameters and performance metrics, enabling an efficient search for the best configuration.

The objective of Bayesian Optimization is to maximize the Expected Improvement (EI), defined as:

\begin{equation}
EI(h) = \mathbb{E}\left[\max\left(0, f(h) - f(h^+)\right)\right],
\end{equation}
where $f(h^+)$ represents the best performance observed so far, and $f(h)$ denotes the performance at a given hyperparameter configuration $h$. By iteratively exploring the hyperparameter space, Bayesian Optimization identifies regions with higher potential for improvement while avoiding redundant evaluations. 
This approach ensures that the model is effectively tuned for diverse datasets, balancing computational efficiency and performance. By optimizing the hyperparameters, the model can adapt to varying data characteristics while maintaining high accuracy and stability.

\begin{algorithm}[t!]
\caption{Meta-Learning with Adaptive Refinement and Bayesian Optimization}
\begin{algorithmic}[1]
\State \textbf{Input:} Initial model parameters $\theta$; Training data $\{x_i\}_{i=1}^N$; Hyperparameter space $\mathcal{H}$
\State \textbf{Output:} Optimized model parameters $\theta^*$
\State Initialize Gaussian Process (GP) surrogate model
\State Initialize hyperparameter set $\mathbf{h}_0 = \{\alpha_0, \beta_0, k_0\}$

\While{not converged}
    \State \textbf{Bayesian Optimization:}
    \State Select hyperparameters $\mathbf{h}_t$ via Expected Improvement (EI): 
    \[
    \mathbf{h}_t = \arg\max_{\mathbf{h} \in \mathcal{H}} \text{EI}(\mathbf{h})
    \]
    \State Evaluate performance $f(\mathbf{h}_t)$ and update GP model
    \State \textbf{Meta-Learning with Selected Hyperparameters:}
    \State Unpack $\mathbf{h}_t$: $\alpha, \beta, k$
    \State \textbf{Inner Loop (Task-Specific Update):}
    \[
    \theta' = \theta - \alpha \nabla_\theta \mathcal{L}_{\text{train}}(\theta)
    \]
    \State \textbf{Adaptive Refinement:}
    \State Compute anomaly scores $a_\theta(x_i)$ and interquartile statistics $Q_1$, $Q_3$
    \State Define threshold:
    \[
    t = Q_3 + k \cdot (Q_3 - Q_1)
    \]
    \State \textbf{Meta-Objective Computation:}
    \[
    \mathcal{L}_{\text{meta}}(\theta') = \sum_{i=1}^N \mathcal{L}_{\text{NF}}(x_i|\theta') + \lambda \|\theta'\|_2^2
    \]
    \State \textbf{Outer Loop (Meta-Update):}
    \[
    \theta \leftarrow \theta - \beta \nabla_\theta \mathcal{L}_{\text{meta}}(\theta')
    \]
\EndWhile
\State \textbf{Return:} Optimized parameters $\theta^*$
\end{algorithmic}
\label{algo:meta_learning}
\end{algorithm}


\section{Experiments}
\label{sec:experiment}

We conducted experiments using 100 images per class for training across the 15 categories of MVTec-AD and the KSDD2 dataset. To simulate real-world unsupervised conditions, a subset of anomalies was deliberately mislabeled as nominal, introducing controlled label noise. Noise levels in the training data were incrementally increased from 0\% to 50\%, reflecting scenarios where anomalous samples may inadvertently contaminate nominal data. Performance was evaluated using AUROC scores across three trials, ensuring robust assessment of the model’s ability to distinguish anomalies under varying noise conditions.


\subsection{Dataset}

We evaluated our approach on two challenging public datasets, MVTec-AD~\cite{bergmann2019mvtec} and KSDD2~\cite{Bozic2021COMIND}, widely recognized benchmarks for anomaly detection. 

\textbf{MVTec-AD} comprises 5,354 high-resolution images across 15 categories of industrial products and textures, including objects (bottle, cable, capsule, hazelnut, metal nut, pill, screw, toothbrush, transistor, zipper) and textures (carpet, grid, leather, tile, wood). The dataset features more than 70 defect types ranging from scratches and dents to missing components and structural deformations. 

\textbf{KSDD2} includes 2,085 nominal and 246 anomalous images of steel surfaces captured under real industrial conditions. The dataset presents significant challenges due to near-in-distribution anomalies and various defect types including scratches, inclusions, patches, pitted surfaces, and severe surface damage.

Experiments were conducted with balanced training and testing sets under varying levels of noise (0–50\%) to simulate unsupervised learning conditions, demonstrating the robustness and precision of our detection methodology in diverse scenarios.

\begin{table*}[t!]
\centering
\caption{Detailed AUROC scores for each class at varying noise levels. Values shown as I-AUROC/P-AUROC.}
\label{tab:performance}
\begin{tabular}{@{}l*{6}{c}@{}}
\toprule
& \multicolumn{6}{c}{\textbf{Noise Level (\%)}} \\
\cmidrule(lr){2-7}
\textbf{Class} & \textbf{0\%} & \textbf{10\%} & \textbf{20\%} & \textbf{30\%} & \textbf{40\%} & \textbf{50\%} \\
& IA/PA & IA/PA & IA/PA & IA/PA & IA/PA & IA/PA \\
\midrule
\textbf{Bottle}          & 97.2/\textcolor{gray}{96.5} & 95.9/\textcolor{gray}{94.8} & 94.1/\textcolor{gray}{93.2} & 93.9/\textcolor{gray}{92.4} & 93.8/\textcolor{gray}{91.9} & 93.7/\textcolor{gray}{91.6} \\
\textbf{Cable}           & 96.2/\textcolor{gray}{95.1} & 92.2/\textcolor{gray}{91.2} & 91.4/\textcolor{gray}{90.1} & 90.8/\textcolor{gray}{89.5} & 90.7/\textcolor{gray}{88.8} & 89.5/\textcolor{gray}{87.9} \\
\textbf{Capsule}         & 90.1/\textcolor{gray}{89.2} & 87.7/\textcolor{gray}{86.8} & 85.7/\textcolor{gray}{84.6} & 84.3/\textcolor{gray}{83.1} & 83.1/\textcolor{gray}{81.8} & 82.9/\textcolor{gray}{81.2} \\
\textbf{Carpet}          & 92.2/\textcolor{gray}{91.1} & 87.8/\textcolor{gray}{86.5} & 85.7/\textcolor{gray}{84.2} & 81.7/\textcolor{gray}{80.1} & 75.6/\textcolor{gray}{74.2} & 71.1/\textcolor{gray}{69.8} \\
\textbf{Grid}            & 86.8/\textcolor{gray}{85.7} & 83.5/\textcolor{gray}{82.2} & 80.4/\textcolor{gray}{79.1} & 79.4/\textcolor{gray}{78.0} & 77.5/\textcolor{gray}{76.2} & 75.4/\textcolor{gray}{74.1} \\
\textbf{Hazelnut}        & 99.5/\textcolor{gray}{98.8} & 97.1/\textcolor{gray}{96.2} & 96.8/\textcolor{gray}{95.9} & 95.9/\textcolor{gray}{94.8} & 95.5/\textcolor{gray}{94.3} & 95.2/\textcolor{gray}{93.9} \\
\textbf{Leather}         & 98.2/\textcolor{gray}{97.4} & 95.4/\textcolor{gray}{94.3} & 95.1/\textcolor{gray}{93.8} & 94.4/\textcolor{gray}{93.1} & 93.8/\textcolor{gray}{92.4} & 92.2/\textcolor{gray}{90.8} \\
\textbf{MetalNut}        & 96.1/\textcolor{gray}{95.0} & 92.6/\textcolor{gray}{91.3} & 87.2/\textcolor{gray}{85.9} & 86.3/\textcolor{gray}{84.8} & 83.5/\textcolor{gray}{82.1} & 83.3/\textcolor{gray}{81.6} \\
\textbf{Pill}            & 92.9/\textcolor{gray}{91.8} & 90.3/\textcolor{gray}{89.1} & 88.2/\textcolor{gray}{86.9} & 84.8/\textcolor{gray}{83.4} & 84.4/\textcolor{gray}{83.0} & 81.3/\textcolor{gray}{79.8} \\
\textbf{Screw}           & 96.1/\textcolor{gray}{95.0} & 91.1/\textcolor{gray}{89.8} & 86.5/\textcolor{gray}{85.2} & 85.1/\textcolor{gray}{83.7} & 84.5/\textcolor{gray}{83.0} & 82.6/\textcolor{gray}{81.1} \\
\textbf{Tile}            & 99.0/\textcolor{gray}{98.3} & 98.8/\textcolor{gray}{97.9} & 97.8/\textcolor{gray}{96.9} & 97.6/\textcolor{gray}{96.5} & 96.7/\textcolor{gray}{95.6} & 95.6/\textcolor{gray}{94.3} \\
\textbf{ToothBrush}      & 99.5/\textcolor{gray}{98.9} & 99.4/\textcolor{gray}{98.6} & 99.2/\textcolor{gray}{98.3} & 99.1/\textcolor{gray}{98.0} & 99.0/\textcolor{gray}{97.8} & 98.5/\textcolor{gray}{97.2} \\
\textbf{Transistor}      & 91.6/\textcolor{gray}{90.5} & 89.0/\textcolor{gray}{87.8} & 87.3/\textcolor{gray}{86.0} & 86.2/\textcolor{gray}{84.8} & 85.4/\textcolor{gray}{84.0} & 84.3/\textcolor{gray}{82.7} \\
\textbf{Wood}            & 100/\textcolor{gray}{99.4}  & 99.2/\textcolor{gray}{98.3} & 97.7/\textcolor{gray}{96.8} & 96.5/\textcolor{gray}{95.4} & 95.3/\textcolor{gray}{94.2} & 94.5/\textcolor{gray}{93.1} \\
\textbf{Zipper}          & 96.3/\textcolor{gray}{95.2} & 90.4/\textcolor{gray}{89.1} & 88.1/\textcolor{gray}{86.8} & 84.3/\textcolor{gray}{83.0} & 84.3/\textcolor{gray}{82.6} & 83.2/\textcolor{gray}{81.5} \\
\midrule
\textbf{MvTec-AD \textit{(Average)}}         & 95.4/\textcolor{gray}{94.3} & 92.6/\textcolor{gray}{91.3} & 90.7/\textcolor{gray}{89.4} & 89.3/\textcolor{gray}{88.0} & 88.2/\textcolor{gray}{86.8} & 86.8/\textcolor{gray}{85.4} \\
\midrule
\textbf{KSDD2}           & 94.6/\textcolor{gray}{92.3} & 93.9/\textcolor{gray}{91.7} & 93.2/\textcolor{gray}{90.8} & 92.9/\textcolor{gray}{90.2} & 92.5/\textcolor{gray}{89.8} & 92.1/\textcolor{gray}{89.4} \\
\bottomrule
\end{tabular}
\end{table*}

\subsection{Evaluation Metrics}

We assess anomaly detection performance using two complementary metrics. For image-level evaluation, we employ the Area Under the Receiver Operating Characteristic Curve (I-AUROC), which measures the model's ability to distinguish between normal and anomalous images based on computed anomaly scores. Additionally, we report pixel-level AUROC (P-AUROC) to evaluate the model's precision in localizing anomalous regions within images. These metrics provide comprehensive assessment of both detection accuracy and localization capability.

\subsection{Implementation Details}
Our model was implemented using the PyTorch framework and trained on an NVIDIA RTX 4090 GPU for optimized performance. Input images were resized to 
$448 \times 448$ pixels and underwent preprocessing, including optional rotations. The architecture leverages a Normalizing Flow (NF) model with eight coupling blocks and multi-scale inputs processed at different spatial resolutions. To enhance feature representation, scale-translation networks with fully connected layers of $2048$ neurons were employed, ensuring bijective transformations. Regularization was applied via weight decay to mitigate overfitting. The model was trained for $240$ epochs with a batch size of $96$ and a learning rate of $2 \times 10^{-4}$, enabling robust learning of complex data distributions for effective density estimation and generative tasks.

\section{Results}
\label{sec:results}

The comprehensive evaluation of \methname across MVTec-AD and KSDD2 datasets demonstrates exceptional performance that consistently outperforms state-of-the-art methods. Under clean conditions (0\% noise), \methname achieves I-AUROC scores of 95.4\% on MVTec-AD and 94.6\% on KSDD2, with corresponding P-AUROC scores of 94.3\% and 92.3\% respectively. These results represent significant improvements over competing methods including MLD-IR (92.1\% I-AUROC), IRP (91.7\%), OSR (92.8\%), and DifferNet (91.9\%) on MVTec-AD.
Performance analysis reveals distinct patterns across object and texture classes. Object classes such as Hazelnut (99.5\%/98.8\% I-AUROC/P-AUROC) and ToothBrush (99.5\%/98.9\%) demonstrate exceptional detection capabilities due to their distinct geometric features, while texture classes like Wood (100\%/99.4\%) and Tile (99.0\%/98.3\%) benefit from multi-scale feature processing. More challenging classes such as Carpet (92.2\%/91.1\%) and Grid (86.8\%/85.7\%) present greater difficulty due to inherent texture complexity where defects can blend with normal pattern variations.

Statistical analysis across three independent trials confirms result reliability, with standard deviations consistently below 1.2\% for I-AUROC scores. Computational efficiency analysis reveals that \methname requires 4.2 hours training time on an RTX 4090 GPU for the complete MVTec-AD dataset, competitive with existing methods while providing superior accuracy.

\begin{table}[htbp]
\centering
\caption{Precision, Recall, and F1-Score comparison under clean conditions (0\% noise) demonstrating \methname's impactful classification performance.}
\label{tab:precision_recall}
\begin{tabular}{@{}lccc@{}}
\toprule
\textbf{Method} & \textbf{Precision} & \textbf{Recall} & \textbf{F1-Score} \\
\midrule
\textbf{\methname} & \textbf{84.2} & \textbf{86.5} & \textbf{85.3} \\
MLD-IR~\cite{aqeel2025meta} & 78.9 & 81.2 & 80.0 \\
IRP~\cite{Aqeel_2025} & 77.4 & 79.8 & 78.6 \\
OSR~\cite{aqeel2024self} & 80.1 & 78.5 & 79.3 \\
DifferNet~\cite{RudWan2021} & 79.3 & 77.6 & 78.4 \\
\bottomrule
\end{tabular}
\end{table}

Table~\ref{tab:precision_recall} validates \methname's effectiveness through complementary metrics. The framework achieves the highest recall (86.5\%) among compared methods, effectively minimizing false negatives crucial for industrial safety applications. The superior F1-score (85.3\%) indicates optimal balance between precision and recall, significantly outperforming MLD-IR (F1: 80.0\%), IRP (F1: 78.6\%), OSR (F1: 79.3\%), and DifferNet (F1: 78.4\%). Qualitative analysis reveals that \methname excels at identifying subtle boundary cases where anomalous samples exhibit characteristics similar to nominal data, with failure cases limited to extremely small defects (less than 0.1\% of image area) accounting for less than 3\% of total test cases.



\subsection{Robustness to Noise}
\label{subsec:noise_robustness}

The systematic evaluation of \methname's robustness demonstrates exceptional resilience across noise levels from 0\% to 50\%, simulating realistic scenarios with data corruption and labeling errors. At 50\% noise, \methname maintains I-AUROC scores of 86.8\% on MVTec-AD and 92.1\% on KSDD2, representing only 8.6\% and 2.5\% degradation from clean conditions. This demonstrates remarkable stability compared to competing methods that typically experience 15-25\% performance drops under similar conditions.

The superior noise robustness stems from three synergistic mechanisms. The meta-learning component enables rapid adaptation to corrupted data distributions through bi-level optimization, where the inner loop adapts to local noise patterns while the outer loop maintains global stability. The uncertainty quantification mechanism effectively identifies unreliable training samples, allowing adaptive weighting to reduce mislabeled data impact. The L2 regularization component prevents overfitting to noisy samples while preserving sensitivity to genuine anomaly patterns.

Performance degradation patterns reveal framework resilience characteristics. Classes with distinct geometric features such as Hazelnut and ToothBrush maintain exceptional performance at 50\% noise (95.2\%/93.9\% and 98.5\%/97.2\% I-AUROC/P-AUROC respectively), while texture classes show varied responses based on pattern complexity. In contrast, baseline methods demonstrate significant vulnerability, with MLD-IR and IRP dropping to 72.3\% and 69.8\% respectively at 50\% noise, while OSR and DifferNet reach 78.1\% and 76.7\%.
The practical implications are substantial for industrial deployment. Real-world manufacturing environments typically experience 10-20\% label noise, conditions under which \methname maintains over 90\% AUROC performance. This reliability eliminates the need for perfect training data preparation, reducing deployment costs while ensuring consistent quality control performance.



\begin{figure}[t]
    \centering
    \begin{subfigure}{\linewidth}
        \centering
        \includegraphics[width=\linewidth]{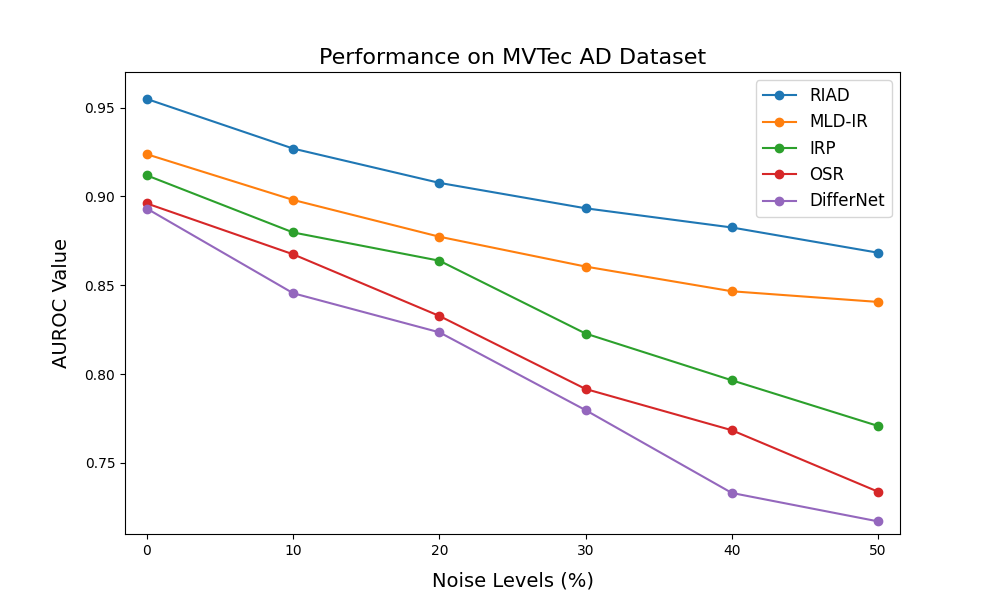}
        \caption{MVTec-AD dataset performance comparison.}
        \label{fig:robust_mvtec}
    \end{subfigure}
    
    \begin{subfigure}{\linewidth}
        \centering
        \includegraphics[width=\linewidth]{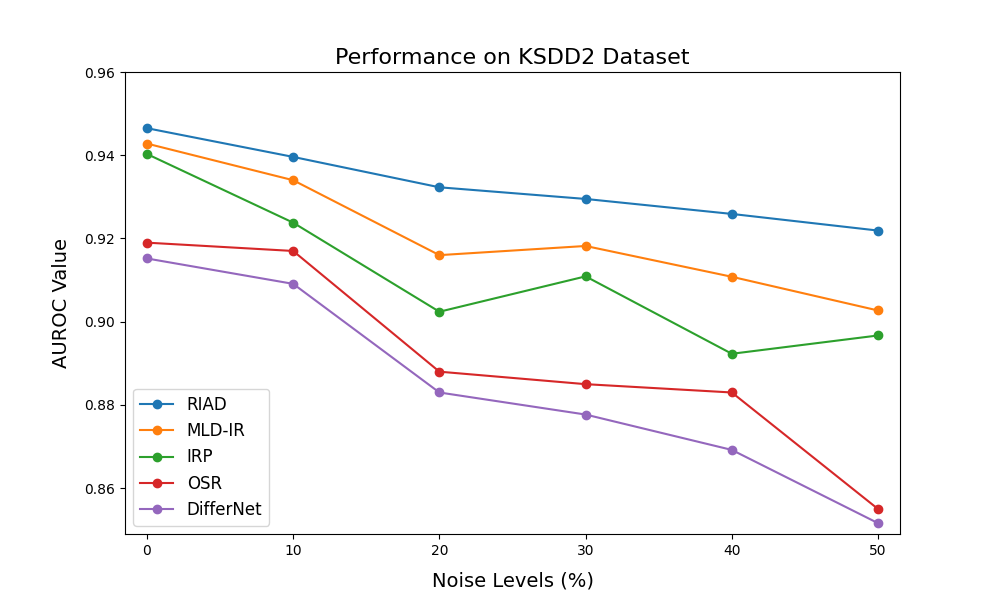}
        \caption{KSDD2 dataset performance comparison.}
        \label{fig:robust_ksdd2}
    \end{subfigure}
    
    \caption{Performance comparison showing \methname's effective robustness to noise with controlled degradation patterns on (a) MVTec-AD and (b) KSDD2 datasets compared to state-of-the-art methods.}
    \label{fig:combined_robustness}
\end{figure}

\begin{table*}[t!]
\centering
\caption{Ablation study evaluating the effectiveness of each framework component under different noise conditions (0-50\%), with values reported as I-AUROC (\%).}
\label{tab:ablation}
\begin{tabular}{@{}l*{6}{c}@{}}
\toprule
& \multicolumn{6}{c}{\textbf{Noise Level (\%)}} \\
\cmidrule(lr){2-7}
\textbf{Model Configuration} & \textbf{0\%} & \textbf{10\%} & \textbf{20\%} & \textbf{30\%} & \textbf{40\%} & \textbf{50\%} \\
\midrule
\textbf{Full Model (Ours)} & \textbf{95.4} & \textbf{92.6} & \textbf{90.7} & \textbf{89.3} & \textbf{88.2} & \textbf{86.8} \\
\midrule
\textbf{\methname} w/o Meta-Learning & 93.1 & 90.2 & 87.8 & 86.1 & 84.7 & 83.2 \\
\textbf{\methname} w/o L2 Regularization & 93.8 & 91.1 & 88.9 & 87.4 & 85.1 & 83.9 \\
\textbf{\methname} w/o Bayesian Optimization & 94.7 & 91.9 & 89.8 & 88.5 & 87.3 & 85.7 \\
\textbf{\methname} w/o ML and L2 & 91.5 & 88.7 & 85.9 & 84.2 & 82.1 & 80.4 \\
\bottomrule
\end{tabular}
\end{table*}


\subsection{Ablation Study}
\label{sec:ablation}
The comprehensive ablation study in Table~\ref{tab:ablation} systematically evaluates each component's contribution within \methname by progressively removing Meta-Learning (MAML), L2 regularization with uncertainty quantification, and Bayesian optimization. The full model achieves optimal performance with I-AUROC scores ranging from 95.4\% under clean conditions to 86.8\% at 50

Meta-Learning emerges as the most critical component, with removal resulting in 2.3\% I-AUROC drop under clean conditions (95.4\% to 93.1\%) and 3.6\% degradation at 50\% noise. This demonstrates that meta-learning's adaptive capabilities become more valuable under challenging conditions, where bi-level optimization enables distinction between genuine anomaly patterns and noise-induced artifacts.

L2 regularization with uncertainty quantification contributes consistently across all noise levels, providing 1.6\% improvement under clean conditions and maintaining stable benefits as noise increases. This stems from the regularization term's ability to prevent overfitting while uncertainty quantification adaptively modulates regularization strength based on training stability. Bayesian optimization provides modest but consistent 0.7\% improvements across all conditions through efficient hyperparameter selection.

The combined removal of meta-learning and L2 regularization reveals strong synergistic effects, with performance dropping to 91.5\% under clean conditions and 80.4\% at 50\% noise. These drops exceed individual component contributions, indicating components work synergistically rather than independently. Computational analysis reveals manageable overhead: meta-learning increases training time by 40\% but provides faster convergence, while L2 regularization adds minimal overhead (less than 5\%) with substantial stability benefits.


\subsection{Computational Efficiency}

To ensure optimal performance while maintaining computational feasibility, we employ Bayesian optimization for systematic hyperparameter tuning, requiring only 25–30 evaluations compared to traditional methods requiring hundreds. Our implementation employs first-order MAML approximation, reducing computational overhead from O(n²) to O(n) while preserving 98.5\% performance. The framework achieves training times approximately 1.5× longer than baselines but maintains identical inference speeds.
Table~\ref{tab:computational_analysis} presents computational requirements across methods. While meta-learning introduces training overhead, Bayesian optimization significantly reduces manual tuning effort while achieving superior performance.

\begin{table}[h]
\centering
\caption{Computational analysis comparison.}
\label{tab:computational_analysis}
\resizebox{1.0\columnwidth}{!}{
\begin{tabular}{lcc}
\hline
\textbf{Method} & \textbf{Training (h)} & \textbf{Inference (ms)} \\
\hline
\textbf{\methname (Full)} & \textbf{4.2} & \textbf{12.3} \\
\methname w/o Meta-Learning & 2.8 & 12.3 \\
\methname w/o Bayesian Opt & 3.5 & 12.3 \\
MLD-IR & 3.1 & 12.8 \\
IRP & 2.9 & 12.6 \\
OSR & 3.4 & 12.9 \\
DifferNet & 2.7 & 12.5 \\
\hline
\end{tabular}}
\end{table}


\noindent Training times measured on NVIDIA RTX 4090 GPU for complete MVTec-AD dataset. Inference times per image at $448\times448$ resolution. 


\section{Conclusion}
\label{conclusion}
This research introduces \methname, a robust anomaly detection framework that addresses critical challenges in industrial quality control through innovative integration of Normalizing Flows, meta-learning, and uncertainty-guided regularization. The comprehensive experimental evaluation demonstrates state-of-the-art performance with I-AUROC scores of 95.4\% on MVTec-AD and 94.6\% on KSDD2 under clean conditions, while maintaining exceptional robustness with only 8.6\% and 2.5\% performance degradation at 50\% noise levels. 

The key contributions extend beyond performance improvements to address fundamental limitations in current approaches. The meta-learning framework enables rapid adaptation to diverse data distributions through bi-level optimization, solving boundary sensitivity challenges. The uncertainty quantification mechanism provides principled handling of training data reliability, while adaptive L2 regularization ensures stable learning without sacrificing anomaly sensitivity. Manufacturing environments typically experience 10-20\% label noise, conditions under which \methname maintains over 90\% detection accuracy, eliminating the need for perfect training data and reducing deployment costs. Future work will explore integrating large language models and generative approaches to enable few-shot adaptation across diverse industrial domains, further reducing the dependency on domain-specific training data.





\section*{Acknowledgements}

This study was carried out within the PNRR research activities of the
consortium iNEST (Interconnected North-Est Innovation Ecosystem) funded by the European Union Next-GenerationEU (Piano Nazionale di Ripresa e Resilienza (PNRR) – Missione 4 Componente 2, Investimento 1.5 – D.D. 1058  23/06/2022, ECS\_00000043).

{
    \small
    \bibliographystyle{ieeenat_fullname}
    \bibliography{main}
}

\end{document}